\documentclass[10pt,twocolumn,letterpaper]{article}

\usepackage[pagenumbers]{iccv} 

\definecolor{iccvblue}{rgb}{0.21,0.49,0.74}
\usepackage[pagebackref,breaklinks,colorlinks,allcolors=iccvblue]{hyperref}
\usepackage{algorithmic}
\usepackage[ruled]{algorithm2e}
\usepackage{tikz}
\usepackage{makecell} 
\def\paperID{14988} 
\def\confName{ICCV}
\def\confYear{2025}

\title{Breaking the Limits of Quantization-Aware Defenses: QADT-R for Robustness Against Patch-Based Adversarial Attacks in QNNs}

\author{Amira Guesmi\\
NYU Abu Dhabi\\
UAE\\
\and
Bassem Ouni\\
Technology Innovation Institute\\
UAE\\
\and
Muhammad Shafique\\
NYU Abu Dhabi\\
UAE\\
}

\begin{document}
\maketitle

\begin{abstract}
Quantized Neural Networks (QNNs) have emerged as a promising solution for reducing model size and computational costs, making them well-suited for deployment in edge and resource-constrained environments. While quantization is known to disrupt gradient propagation and enhance robustness against pixel-level adversarial attacks, its effectiveness against patch-based adversarial attacks remains largely unexplored. In this work, we demonstrate that adversarial patches remain highly transferable across quantized models, achieving over 70\% attack success rates (ASR) even at extreme bit-width reductions (e.g., 2-bit). This challenges the common assumption that quantization inherently mitigates adversarial threats.
To address this, we propose Quantization-Aware Defense Training with Randomization (QADT-R), a novel defense strategy that integrates Adaptive Quantization-Aware Patch Generation (A-QAPA), Dynamic Bit-Width Training (DBWT), and Gradient-Inconsistent Regularization (GIR) to enhance resilience against highly transferable patch-based attacks. A-QAPA generates adversarial patches within quantized models, ensuring robustness across different bit-widths. DBWT introduces bit-width cycling during training to prevent overfitting to a specific quantization setting, while GIR injects controlled gradient perturbations to disrupt adversarial optimization.
Extensive evaluations on CIFAR-10 and ImageNet show that QADT-R reduces ASR by up to 25\% compared to prior defenses such as PBAT and DWQ. Our findings further reveal that PBAT-trained models, while effective against seen patch configurations, fail to generalize to unseen patches due to quantization shift. Additionally, our empirical analysis of gradient alignment, spatial sensitivity, and patch visibility provides insights into the mechanisms that contribute to the high transferability of patch-based attacks in QNNs. 
\end{abstract}
    
\section{Introduction}
\label{sec:intro}

Deep learning models are widely deployed in real-world applications, including autonomous systems, edge AI, and mobile computing, where computational efficiency and robustness are critical. Quantized Neural Networks (QNNs) have emerged as an effective solution to reduce model size and computational overhead while maintaining competitive accuracy \cite{liu2021flexi, katare2023survey, zhang2021medq, tonellotto2021neural, hernandez2024optimizing}. While quantization enhances robustness against pixel-level adversarial perturbations by disrupting gradient propagation, its impact on structured attacks, such as patch-based adversarial attacks, remains largely unexplored.

Most existing adversarial robustness studies focus on pixel-level perturbations, demonstrating that low-bit quantization disrupts gradients, making pixel-based attacks less effective \cite{Li2024InvestigatingTI, yang2024quantization}. However, patch-based adversarial attacks operate differently, introducing high-contrast, spatially constrained modifications that persist across architectures and quantization settings. Unlike pixel-level attacks, which rely on subtle, imperceptible changes, adversarial patches induce dominant feature activations that remain effective even at extreme quantization levels (e.g., 2-bit, 4-bit), highlighting a major vulnerability in QNNs.

To mitigate adversarial threats, prior defenses have explored adversarial training and quantization-aware techniques. Standard Adversarial Training (SAT) \cite{madry2017towards} improves robustness against pixel-based attacks, but struggles against structured perturbations such as adversarial patches. Patch-Based Adversarial Training (PBAT) \cite{rao2020adversarial} aims to improve resilience by incorporating adversarial patches during training, but suffers from overfitting to seen patch configurations and fails under unseen quantization settings due to quantization shift. Double-Win Quantization (DWQ) \cite{fu2021double}, a random precision training method, enhances robustness against pixel-based perturbations but remains vulnerable to patch-based adversaries.

To address these limitations, we propose Quantization-Aware Defense Training with Randomization (QADT-R), a novel adversarial defense designed to neutralize the transferability of patch-based attacks in QNNs. QADT-R integrates three key components: \textit{Adaptive Quantization-Aware Patch Generation (A-QAPA)}, which ensures adversarial patches remain effective across quantization settings; \textit{Dynamic Bit-Width Training (DBWT)}, which prevents overfitting to a single precision by cycling through multiple bit-widths during training; and \textit{Gradient-Inconsistent Regularization (GIR)}, which disrupts adversarial optimization by introducing random perturbations into the gradient computation. 

\textbf{Our key contributions are:}
\begin{itemize} 
    \item We present the first comprehensive study on the transferability of patch-based adversarial attacks across quantization levels and architectures, showing that adversarial patches maintain high attack success rates (ASR $>$ 70\%) even at extreme low-bit precision (e.g., 2-bit).
    \item We propose QADT-R, the first quantization-aware defense strategy specifically designed to counter patch-based attacks. QADT-R integrates adaptive quantization-aware patch generation (A-QAPA), dynamic bit-width training (DBWT), and gradient-inconsistent regularization (GIR) to enhance robustness against highly transferable adversarial patches in QNNs.   
    \item We evaluate QADT-R on CIFAR-10 and ImageNet, demonstrating that it reduces ASR by over 25\% compared to state-of-the-art defenses (PBAT, DWQ), and over 40\% against unseen attack configurations, while maintaining high clean accuracy with a drop of less than 2\%.   
    \item We conduct an ablation study to validate the contribution of each component of QADT-R under both seen and unseen patch configurations, confirming the necessity of A-QAPA, DBWT, and GIR in achieving robust defenses.   
    \item We provide in-depth analyses of gradient alignment, spatial sensitivity, and patch persistence to provide insights into the mechanisms that contribute to the high transferability of patch-based attacks in QNNs.
\end{itemize}

Through these contributions, we establish QADT-R as a scalable and adaptive defense mechanism, significantly enhancing the robustness of quantized deep learning models against structured adversarial threats.

\section{Related Work}
\label{sec:related_work}
Most adversarial attack studies on QNNs focus on gradient-based pixel-level perturbations, showing that lower-bit quantization weakens attack transferability by obscuring gradient signals \cite{Li2024InvestigatingTI, yang2024quantization}. However, patch-based adversarial attacks \cite{lavan, googleap, chen2022shape} operate differently by introducing high-contrast, localized perturbations that remain highly transferable across architectures and quantization levels. 
Patch-based attacks exploit localized adversarial patterns, making them resilient to common defenses. Patch-Based Adversarial Training (PBAT) \cite{rao2020adversarial} partially mitigates these threats by incorporating adversarial patches into the training process. However, PBAT remains vulnerable in quantized models due to the quantization shift, which disrupts adversarial feature alignment.

Prior works have explored various techniques to improve adversarial robustness in QNNs:
Standard Adversarial Training (SAT): Traditional adversarial training strategies (e.g., PGD-based AT \cite{madry2017towards}) are highly effective against pixel-level attacks but fail against structured perturbations such as adversarial patches.
Double-Win Quantization (DWQ) \cite{fu2021double}: A random precision training (RPT) method that enhances robustness against pixel-based adversaries, but remains vulnerable to patch-based attacks due to limited adaptability to spatial perturbations.
Regularization-Based Defenses: Methods such as feature-space regularization \cite{song2020improving} and gradient obfuscation techniques \cite{athalye2018obfuscated} attempt to disrupt adversarial feature alignment but do not generalize well to patch-based attacks.
Ensemble-Based Defenses: Approaches such as stochastic bit-width inference \cite{sen2020empir} improve adversarial robustness in QNNs but offer little protection against highly localized perturbations like adversarial patches.
Table~\ref{tab:qadt_r_comparison} summarizes the key differences between QADT-R and existing defenses.

\begin{table}[ht]
    \centering
    \footnotesize  
    \renewcommand{\arraystretch}{0.7}  
    \setlength{\tabcolsep}{2pt}  
    \begin{tabular}{|c|c|c|c|}
        \hline
        \textbf{Defense} & \makecell{\textbf{Training} \\\textbf{Strategy}} & \makecell{\textbf{Inference} \\ \textbf{Strategy}} & \makecell{\textbf{Attack} \\\textbf{Resistance}} \\ \hline
        AT \cite{madry2017towards} & \makecell{Fixed-bit \\ Adversarial \\ Training} & \makecell{Fixed-bit \\ inference} & \makecell{Weak against \\ patch-based \\ attacks} \\ \hline
        PBAT \cite{rao2020adversarial} & \makecell{Patch-based \\ Adversarial \\ Training} & \makecell{Fixed-bit \\ inference} & \makecell{Overfits to \\ patch size, \\ location, and \\ bit-width, \\ vulnerable to \\ unseen attacks} \\ \hline
        DWQ~\cite{fu2021double} & \makecell{Random \\ Precision \\ Training (RPT)} & \makecell{Global \\Random \\ Precision \\ Inference \\(RPI)} & \makecell{Strong against \\ pixel-level \\ perturbations, \\ weak against \\ patch-based \\ attacks} \\ \hline
        \makecell{\textbf{QADT-R} \\ \textbf{(Ours)}} & \makecell{Adaptive Quantization-\\Aware Patch Generation \\ (A-QAPA) + \\ Dynamic Bit-Width \\ Training (DBWT) + \\ Gradient-Inconsistent \\ Regularization (GIR)} & \makecell{Bit-Width-\\Invariant \\ Inference} & \makecell{Strong against \\ patch-based \\ attacks \\ across unseen \\ bit-widths and \\ configurations} \\ \hline
    \end{tabular}
    \caption{Comparison between QADT-R and Existing Defenses.}
    \label{tab:qadt_r_comparison}
\end{table}

\section{Methodology}
\label{methodology}

In this section, we present our key observations: (1) Low-bit quantization weakens pixel-level adversarial attacks but does not provide significant robustness against patch-based attacks, and (2) Patch-based adversarial attacks remain highly transferable across different quantization bit-widths and architectures. Next, we analyze the limitations of existing adversarial defenses that fail to mitigate structured patch-based attacks. Finally, we introduce QADT-R as a countermeasure, leveraging bit-width randomization and quantization-aware adversarial training to mitigate these vulnerabilities.
\subsection{Preliminaries: Patch-based Attacks against QNNs}
Existing research suggests that quantization weakens pixel-level perturbations due to gradient distortion. However, our findings indicate that patch-based attacks remain highly effective across different quantization levels and architectures, making them a critical security concern for real-world QNN deployment.

\noindent\textbf{Observation 1: Low-bit quantization weakens pixel-level adversarial attacks.}
To evaluate the effect of quantization on adversarial robustness, we assess the resilience of quantized models against pixel-level adversarial attacks. Specifically, we consider FGSM, PGD, MIFGSM, and AutoAttack, which are commonly used gradient-based attacks to assess adversarial vulnerability.

\begin{figure*}[t] \centering \includegraphics[width=\linewidth]{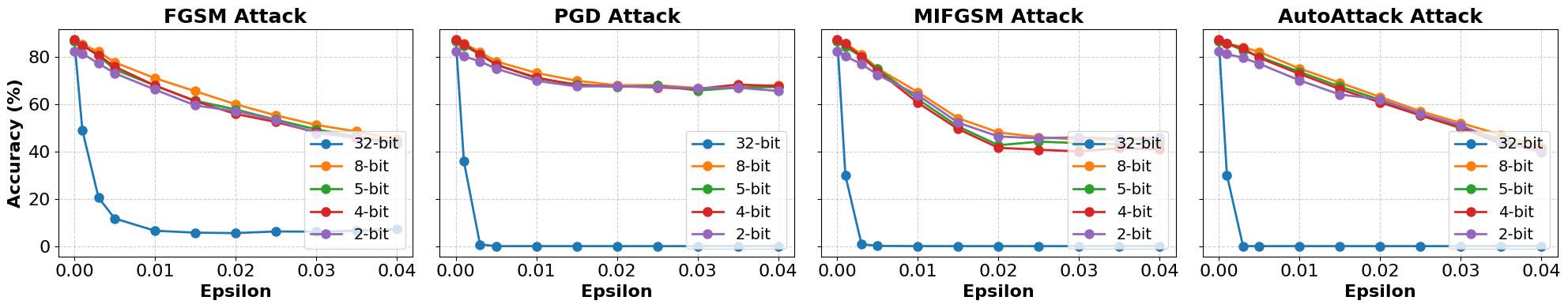} \caption{Model accuracy for ImageNet ResNet-34 across different quantization levels under various pixel-level adversarial attacks.} \label{fig:pixel} \end{figure*}

Figure~\ref{fig:pixel} presents the classification accuracy of ResNet-34 models trained on ImageNet at different quantization levels under various adversarial perturbations.
At low $\epsilon$ values, quantized models (especially 2-bit and 4-bit) maintain higher accuracy compared to their full-precision (32-bit) counterparts, indicating that quantization can mitigate small adversarial perturbations. Despite increased robustness at lower $\epsilon$, low-bit models exhibit a steeper decline in accuracy as attack strength increases, suggesting that adversarial resilience decreases under stronger perturbations. Reduced precision introduces gradient distortion and numerical instability, making pixel-wise attacks less effective in low-bit models. 

These findings align with previous research showing that low-bit quantization disrupts adversarial gradients, reducing the effectiveness of small-scale perturbations. However, this phenomenon applies only to pixel-level perturbations. As we discuss in Observation 2, patch-based attacks do not exhibit the same vulnerability to quantization, making them a persistent security threat even in low-bit QNNs.
\noindent\textbf{Observation 2: Low-bit quantization does not significantly weaken patch-based adversarial attacks.}
To evaluate the impact of quantization on patch-based adversarial attacks, we generate adversarial patches using a 32-bit model and transfer these patches to models quantized to different bit-widths (8-bit, 4-bit, 2-bit). The evaluation includes both QAT and PTQ settings. We measure the attack success rate (ASR) at each quantization level to analyze whether quantization reduces the effectiveness of structured adversarial patches.

\noindent\textbf{i) Quantization-Aware Training (QAT) Results:} 
\begin{table}[ht]
    \centering
    \footnotesize  
    \renewcommand{\arraystretch}{0.9}  
    \setlength{\tabcolsep}{3pt}  
    \begin{tabular}{|c|c|c|c|c||c|c|c|c|c|}
    \hline 
        \textbf{Attack}  & \multicolumn{4}{c||}{\textbf{LAVAN}} &\multicolumn{4}{c|}{\textbf{GAP}} \\
          \hline
      \textbf{Model} &    \textbf{32-bit} & \textbf{8-bit}  & \textbf{4-bit} & \textbf{2-bit} &    \textbf{32-bit} & \textbf{8-bit}  & \textbf{4-bit} & \textbf{2-bit}\\
    \hline
      Res-56  & 86.43  & 83.24   & 76.22 & 73.08 & 84.40  & 56.69   & 54.22  & 47.91 \\ \hline
      Res-20  & 87.22  & 83.73   & 77.30 & 74.18 & 84.71  & 59.61  & 58.45 &  50.31\\ \hline
      VGG-19  & 88.95  & 85.56   & 79.81 & 77.19 &  95.79 & 59.65  & 48.7 & 40.69 \\ \hline
      VGG-16  & 87.17  & 84.73   & 78.29 & 76.67  & 95.71 & 64.24 & 52.04  & 48.90  \\ 
    \hline
    \end{tabular}
    \caption{ASR (\%) of LAVAN and GAP attacks (6x6 patch) across different QNNs on CIFAR-10.}
    \label{tab:experiment2_lavan}
\end{table}
As shown in Table~\ref{tab:experiment2_lavan}, adversarial patches maintain high attack success rates across all bit-widths. Even at the lowest bit-width (2-bit), ASR remains above 70\%, demonstrating that structured adversarial patches remain highly effective despite precision reduction.

\begin{table}[ht]
    \centering
    \footnotesize  
    \renewcommand{\arraystretch}{0.9}  
    \setlength{\tabcolsep}{3pt}  
    \begin{tabular}{|c|c|c|c|c||c|c|c|c|}
        \hline
         & \multicolumn{4}{c||}{\textbf{ResNet34}} & \multicolumn{4}{c|}{\textbf{ResNet18}} \\ 
        \cline{2-9}
        \textbf{NP} & \textbf{32-bit} & \textbf{5-bit} &  \textbf{4-bit} &\textbf{2-bit} & \textbf{32-bit}  & \textbf{5-bit} & \textbf{4-bit} & \textbf{2-bit} \\ \hline
        0.1 & 99.31 & 66.32 & 63.56 & 56.31 & 99.98 &  72.63 & 67.89 & 65.76 \\ \hline
        0.08 & 98.08 & 64.91 & 59.97  & 52.25 & 99.93 &  66.37& 61.11 & 55.42 \\ \hline
        0.06 & 97.12 & 64.79 & 57.31  & 50.43 & 96.01 & 58.20 & 53.59 & 51.84 \\ \hline
    \end{tabular}
    \caption{ASR (\%) of LAVAN attack with different noise percentages (NP) across different QNNs on ImageNet.}
    \label{tab:experiment2_lavan_resnet34}
\end{table}
Table~\ref{tab:experiment2_lavan_resnet34} further reinforces these findings, showing that even at 2-bit precision, patch-based attacks maintain high attack success rates (above 50\%), confirming that quantization alone is insufficient as a defense mechanism against patch-based adversarial attacks.

\noindent\textbf{ii) Dynamic Quantization (DQ) Results:}
To assess the robustness under DQ, we evaluate attack success rates on dynamically quantized 8-bit models.

\begin{table}[ht]
    \centering
    \footnotesize  
    \renewcommand{\arraystretch}{0.9}  
    \setlength{\tabcolsep}{3.5pt}  
    \begin{tabular}{|c|c|c|c|c|c|c|c|c|}
        \hline
        \textbf{Model} & \multicolumn{2}{c|}{ \textbf{ResNet-56} }& \multicolumn{2}{c|}{ \textbf{ResNet-20}} & \multicolumn{2}{c|}{ \textbf{VGG-19}} &\multicolumn{2}{c|}{ \textbf{VGG-16}} \\ \hline
        BW & 32-bit & 8-bit & 32-bit & 8-bit & 32-bit & 8-bit & 32-bit & 8-bit  \\ \hline
        LAVAN & 86.43 & 84.03 & 87.22 & 83.29 & 88.95 & 76.33 & 87.17 & 71.58 \\ \hline
        GAP   & 84.40 & 82.40 & 84.71 & 53.76 & 95.71& 54.12 & 95.79 & 41.78 \\ \hline
    \end{tabular}
    \caption{ASR (\%) of LAVAN and GAP attacks (6x6 patches) across dynamically quantized models (8-bit) on CIFAR-10.}
    \label{tab:dyn_ptq}
\end{table}
Despite dynamic quantization, Table~\ref{tab:dyn_ptq} shows that adversarial patches remain highly transferable across different models, further confirming the persistence of patch-based attack vulnerabilities.

\noindent\textbf{iii) Post-Training Quantization (PTQ) Results:}

\begin{table}[ht]
    \centering
    \footnotesize  
    \renewcommand{\arraystretch}{0.9} 
    \setlength{\tabcolsep}{4pt} 
    \begin{tabular}{|c|c|c|c|c|c|}
        \hline
             &  \textbf{Attacks}   &   \multicolumn{2}{c|}{ \textbf{LAVAN} } & \multicolumn{2}{c|}{ \textbf{GAP} } \\ \hline
        \textbf{Model} & \textbf{Calibration} & \textbf{32-bit} & \textbf{8-bit} & \textbf{32-bit} & \textbf{8-bit} \\ \hline
        Swin-S & MinMax \cite{LinZSLZ22}& 91.80 & 62.11 & 85.32 & 59.84\\ \hline 
        Swin-S & Percentile \cite{LinZSLZ22}& 93.10 & 63.72 & 87.19 & 61.23\\ \hline 
        DeiT-B & MinMax \cite{LinZSLZ22}& 93.63 & 64.76 & 88.03  & 62.44\\ \hline 
        DeiT-B & Percentile \cite{LinZSLZ22}& 90.12 & 61.51 & 84.37 & 58.73\\ \hline 
    \end{tabular}
    \caption{ASR (\%) of LAVAN and GAP attacks with a noise percentage 0.08 on ImageNet.}
    \label{tab:experiment2_transformer}
\end{table}

Even in PTQ-based Transformer models (Swin-S, DeiT-B), adversarial patches retain high ASR (see Table \ref{tab:experiment2_transformer}), demonstrating that patch-based threats extend beyond CNNs.

\noindent\textbf{Observation 3: Patch-based adversarial attacks exhibit high transferability across architectures.}
Patch-based adversarial attacks not only persist across different quantization bit-widths but also exhibit strong cross-architecture transferability, making them a severe security threat in real-world black-box settings. To evaluate this phenomenon, we generate adversarial patches on a base architecture (e.g., ResNet-20 at 32-bit precision) and transfer them to models with different architectures (e.g., ResNet-56, VGG-16, and VGG-19) trained with QAT at various bit-widths (8-bit, 5-bit, 4-bit, and 2-bit). The Attack Success Rate is recorded for each architecture-bitwidth combination to assess the transferability of adversarial patches across both architectural and quantization changes.

\begin{table}[ht]
    \centering
    \footnotesize  
    \renewcommand{\arraystretch}{0.9}  
    \setlength{\tabcolsep}{3pt}  
    \begin{tabular}{|c|c|c|c|c|c|c|c|c|c|}
    \hline
          &   \multicolumn{4}{c|}{ \textbf{ResNet56}} &  \multicolumn{4}{c|}{\textbf{VGG-19}}  \\ \hline
     Bitwidth      & 32-bit & 8-bit & 4-bit & 2-bit & 32-bit & 8-bit & 4-bit & 2-bit \\
    \hline                 
      \textbf{ResNet20} &     84.17 & 79.62  & 77.66 & 75.21  &  78.82 & 74.53 & 72.15 & 70.21 \\
    \hline
          & \multicolumn{4}{c|}{ \textbf{ResNet20} }& \multicolumn{4}{c|}{  \textbf{VGG-19}}  \\ \hline
     Bitwidth      & 32-bit & 8-bit & 4-bit & 2-bit & 32-bit & 8-bit & 4-bit & 2-bit \\
          \hline      
      \textbf{ResNet56} &  84.11  & 77.67  & 75.33 & 71.76  & 77.43 & 75.09  & 73.82 & 71.22\\
    \hline
           & \multicolumn{4}{c|}{  \textbf{ResNet20} }& \multicolumn{4}{c|}{  \textbf{VGG-16}} \\ \hline
     Bitwidth      & 32-bit & 8-bit & 4-bit & 2-bit & 32-bit & 8-bit & 4-bit & 2-bit \\
          \hline     
      \textbf{VGG-19} &  83.23  & 80.87  & 78.11 &  75.32 & 85.32 & 80.42  & 78.23 & 76.44\\
    \hline
            & \multicolumn{4}{c|}{ \textbf{VGG-19}} &  \multicolumn{4}{c|}{  \textbf{ResNet56} } \\ \hline
     Bitwidth     & 32-bit & 8-bit & 4-bit & 2-bit & 32-bit & 8-bit & 4-bit & 2-bit \\
          \hline     
      \textbf{VGG-16}  & 83.29 & 78.48 & 76.65 &  74.39 & 80.55 &  78.93  & 75.34  & 73.87 \\
    \hline
    \end{tabular}
    \caption{ASR (\%) transfer across different QNNs with different bitwidths and architectures on CIFAR-10. }
    \label{tab:experimet5}
\end{table}

As presented in Table~\ref{tab:experimet5}, a patch generated on ResNet-20 achieves an 84.17\% attack success rate on 32-bit ResNet-56 and 78.82\% on 32-bit VGG-19. Patches created on VGG-19 and VGG-16 maintain high success rates when tested on ResNet architectures, confirming their strong cross-architecture transferability. Even at low-bit settings (e.g., 2-bit), patches retain attack success rates above 70\%, highlighting their resilience under quantization-induced transformations.

\subsection{Why Patch-Based Attacks Transfer Well?}
Unlike pixel-level adversarial perturbations, which are highly sensitive to architectural variations and feature alignment, patch-based attacks exploit high-contrast, localized disruptions that remain effective across diverse model structures and quantization settings. Our analysis identifies key factors that contribute to their strong transferability, making them a persistent threat even in quantized models.
For a detailed breakdown of our feature map analysis and additional insights, refer to Section \ref{sec:why} in the supplementary material.

\noindent\textit{\textbf{Localized Perturbations Preserve Adversarial Features:}}
Patch-based attacks modify a concentrated, high-contrast region, which is consistently learned across different architectures. This differs from distributed perturbations in pixel-wise attacks, which are more dependent on architecture-specific features.

\noindent\textit{\textbf{Cross-Architecture Generalization of Feature Representations:}}
Deep neural networks share common hierarchical features extraction properties. Since patches target intermediate- and high-level features, their adversarial effects persist across different architectures.

\noindent\textit{\textbf{Limited Gradient Masking in Patch-Based Attacks:}}
Pixel-wise perturbations often suffer from gradient misalignment and masking effects in quantized models. Patch-based attacks, however, are not gradient-sensitive, as they induce high-contrast, recognizable distortions that remain effective across different model structures.

\noindent\textbf{Observation 4: Standard Adversarial Training and Traditional Patch-Based Adversarial Training Fail in QNNs.}
In this section, we systematically evaluate the limitations of both standard AT \cite{madry2017towards} and PBAT \cite{rao2020adversarial}, highlighting their vulnerabilities against structured adversarial attacks in quantization context.
We assess adversarial robustness across nine primary model configurations and training paradigms including combinations of: Full-Precision, QAT, PTQ, AT and PBAT.

We evaluate the effectiveness of these defenses against two different attack:
pixel-level attack (PGD-20) and patch-based attack (LAVAN with patch size (8x8)). 
As illustrated in Table~\ref{tab:experiment4}, QAT-trained models show higher accuracy retention under attack than PTQ models across all bit-widths. However, QAT models still degrade significantly at higher attack strengths, making adversarial training less effective, but not completely useless.
QAT allows the model to adapt to quantized gradients during training, ensuring that adversarial training is effective within the quantized domain. In contrast, PTQ applies quantization post-training, distorting gradients and reducing robustness against adversarial attacks.
\begin{table}[ht]
    \centering
    \footnotesize  
    \renewcommand{\arraystretch}{0.9}  
    \setlength{\tabcolsep}{4pt}  
    \begin{tabular}{|c|c|c|c|c|c|c|}
    \hline
     & \multicolumn{3}{c|}{  \textbf{Standard Training} } & \multicolumn{3}{c|}{  \textbf{AT-PGD7} } \\ \hline
    \textbf{Epsilon}   &  \textbf{FP} &  \textbf{QAT}  &  \textbf{PTQ}  &  \textbf{FP-AT}  &  \textbf{QAT-AT}  &  \textbf{PTQ-AT}  \\ \hline
     \textbf{0}     & 79.28  & 74.92   & 71.77   &  78.25  & 71.83   &  65.29  \\ \hline
     \textbf{0.015} & 8.02   &  58.63  & 15.53   & 75.3   & 69.27   &  63.46  \\ \hline
     \textbf{0.03}  & 3.21   &  52.64  &  8.32  & 74.53   &  68.6  &  62.85  \\ \hline
    \end{tabular}
    \caption{Model accuracy (\%) under PGD-20 attack across different training techniques and quantization strategies.}
    \label{tab:experiment4}
\end{table}
As shown in Table \ref{tab:experiment4_patch_quant}, PBAT achieves lower ASR for patches with the same bit-width as seen in training, confirming its effectiveness for those specific configurations. When tested on unseen patch quantization levels (4-bit, 2-bit), ASR increases by up to 20\%, revealing that PBAT does not generalize well across varying bit-widths. QAT-PBAT models particularly struggle against low-bitwidth patches, showing that quantization shift affects learned adversarial robustness.
PTQ-PBAT exhibits the highest ASR increase for unseen bit-widths, suggesting post-training quantization does not adequately retain robustness against adversarial patches.
\begin{table}[ht]
    \centering
    \footnotesize  
    \renewcommand{\arraystretch}{0.6}  
    \setlength{\tabcolsep}{4pt}  
    \begin{tabular}{|c|c|c|c|c|c|c|}
    \hline
     & \multicolumn{3}{c|}{  \textbf{Standard Training} } & \multicolumn{3}{c|}{  \textbf{PBAT} } \\ \hline
    \textbf{Size}   &  \textbf{FP} &  \textbf{QAT}  &  \textbf{PTQ}  &  \makecell{\textbf{FP-} \\\textbf{PBAT}}  &  \makecell{\textbf{QAT-} \\ \textbf{PBAT}}  &  \makecell{\textbf{PTQ-} \\ \textbf{PBAT}}  \\ \hline
     \textbf{8x8}    & 88.17  & 81.56   & 85.24   &  40.39  &  40.56  &  45.44  \\ \hline
     \textbf{10x10}    & 92.33  & 84.33   & 87.48  &  57.86  &  56.77  &  60.60  \\ \hline
     \makecell{\textbf{8x8}\\ \textbf{(Unseen 4-bit} \\ \textbf{Patch)}}    & 89.92  & 83.40  & 86.78  &  62.10  &  71.42  &  75.16  \\ \hline
     \makecell{\textbf{10x10}\\ \textbf{(Unseen 2-bit} \\ \textbf{Patch)}} & 91.18  & 85.62  & 87.91  &  65.30  &  78.34  &  81.09  \\ \hline
    \end{tabular}
    \caption{ASR (\%) of LAVAN attack across different training techniques and quantization strategies.}
    \label{tab:experiment4_patch_quant}
\end{table}
\subsection{Quantization-Aware Defense Training with Randomized Bit-Widths (QADT-R)}

\begin{figure}[t] \centering \includegraphics[width=\linewidth]{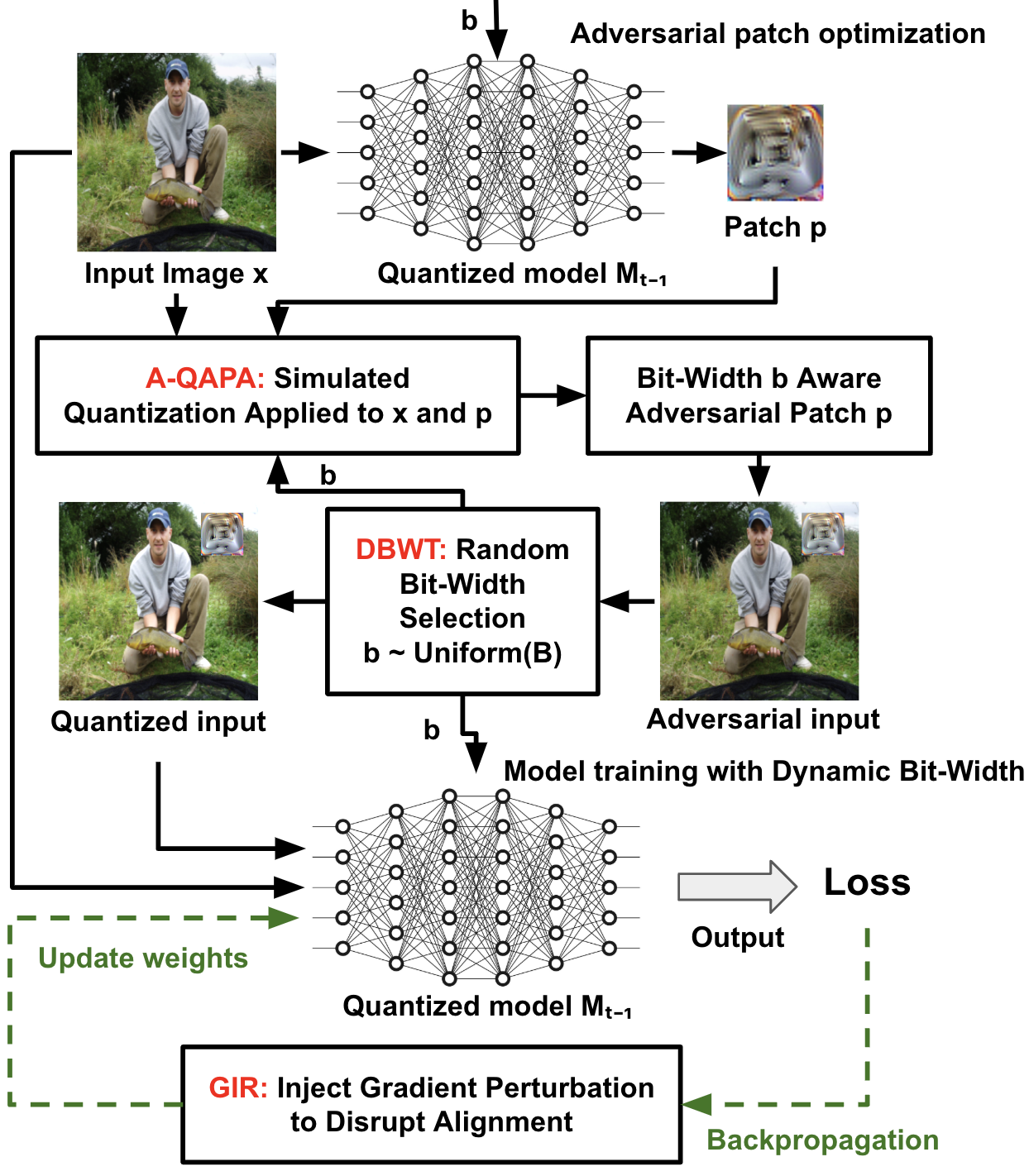} \caption{Overview of Our proposed methodology QADT-R to enhance robustness against patch-based adversarial attacks in QNNs.} \label{fig:methodology} \end{figure}

To mitigate the vulnerability of QNNs to patch-based adversarial attacks, we introduce Quantization-Aware Defense Training with Randomized Bit-Widths (QADT-R). As illustrated in Figure \ref{fig:methodology}, QADT-R enhances model robustness through a three-step process that integrates adaptive adversarial patch generation, dynamic training strategies, and gradient-based regularization to disrupt adversarial optimization. 

\subsubsection{Adaptive Quantization-Aware Patch Generation (A-QAPA)}

Adversarial training for QNNs requires adversarial patches that accurately reflect real-world quantized deployment conditions. Traditional patch-based adversarial training, such as PBAT~\cite{rao2020adversarial}, suffers from overfitting to specific patch configurations and fixed-bit training, making it ineffective against unseen patch perturbations and dynamic quantization settings.

To address these limitations, we introduce Adaptive Quantization-Aware Patch Generation (A-QAPA), which generates adversarial patches under simulated quantization effects. Unlike conventional adversarial patch generation techniques that optimize perturbations on full-precision models, A-QAPA directly integrates bit-width-aware perturbations into adversarial training, ensuring transferability across multiple quantization levels.

A-QAPA ensures that adversarial patches remain effective across different bit-widths by generating them directly within quantized models. It integrates bit-width-aware patch generation, ensuring perturbations align with quantized inference, and patch-aware simulated quantization, introducing randomized quantization effects to improve transferability. Additionally, patch-specific masking applies adaptive quantization only to the adversarial region, preserving clean image areas and preventing unnecessary distortions. This approach enhances robustness, ensuring that patches retain their adversarial properties across varying precision levels.

Given an input image \( x \), an adversarial patch \( p \), and its corresponding patch mask \( M \), the adversarial training sample is constructed as:

\begin{equation}
    x_{adv} = \mathcal{Q}_b(x) \odot (1 - M) + \mathcal{Q}_b(p) \odot M
\end{equation}

where \( \mathcal{Q}_b(\cdot) \) is the quantization function applied at bit-width \( b \), ensuring the patch aligns with real deployment constraints. \( M \) is a binary mask indicating the patch location. \( x \) remains dynamically quantized to match training-time bit-width variations.


Unlike PBAT, which applies adversarial patches at fixed bit-widths, A-QAPA generates adversarial patches under multiple simulated quantization levels, preventing overfitting to a specific bit-width. This ensures that QADT-R can defend against patch attacks at any precision setting, including unseen configurations.
A-QAPA is applied during Dynamic Bit-Width Training (DBWT) to expose the model to a wide range of bit-width-adaptive perturbations, reinforcing generalization against patch-based threats.

\subsubsection{Dynamic Bit-Width Training (DBWT)}

Adversarially training a QNN with patches at a fixed bit-width leads to overfitting, as the model becomes highly specialized in defending against attacks at that particular precision. However, real-world deployment often involves varying bit-width settings (e.g., dynamic quantization, mixed-precision inference). 
To address this, we introduce Dynamic Bit-Width Training (DBWT), where the bit-width is dynamically sampled at each training step. This forces the network to learn bit-width-invariant features, improving robustness across quantized settings.

At each training step, the bit-width \( b \) is randomly sampled from a set of possible bit-widths \( B = \{32, 8, 5, 4, 2\} \). The weights and activations are quantized dynamically to prevent overfitting to a single precision level. Bit-width cycling ensures that the model encounters adversarial patches across multiple quantization levels.

At each training iteration, the bit-width \( (b_w, b_a) \) for weights and activations is sampled from:

\begin{equation}
    (b_w, b_a) \sim \text{Uniform}(B)
\end{equation}

During forward propagation, the model is quantized accordingly:

\begin{equation}
    x_{q} = \mathcal{Q}_{b_w, b_a}(x)
\end{equation}

where \( \mathcal{Q}_{b_w, b_a}(\cdot) \) represents the quantization function at bit-width \( (b_w, b_a) \).

By cycling through multiple bit-widths, the model learns to generalize defenses across varying quantization settings, reducing gradient mismatches and quantization shift effects.

\subsubsection{Gradient-Inconsistent Regularization (GIR)}

While DBWT ensures robustness across multiple bit-widths, the model may still rely on gradient-aligned features that attackers can exploit. To further disrupt adversarial learning, we introduce Gradient-Inconsistent Regularization (GIR).

During backpropagation, the gradient alignment between bit-widths is randomized. Gradient perturbations are injected at lower-bit models to prevent overfitting to a specific bit-width. The goal is to create inconsistencies in the attacker's ability to optimize adversarial patches across different quantization levels.

Mathematically, given the loss function \( L \) and gradients \( \nabla L_{b} \) at bit-width \( b \), we apply gradient perturbation:

\begin{equation}
    \tilde{\nabla L}_{b} = \nabla L_{b} + \alpha \cdot \mathcal{R}(\nabla L_{b})
\end{equation}

where \( \alpha \) is a hyperparameter controlling the magnitude of perturbation. \( R(\nabla L_{b}) \) introduces stochastic perturbation to misalign gradients between bit-widths:
\begin{equation}
    R(\nabla L_{b}) = \beta \cdot \frac{\nabla L_{b}}{\|\nabla L_{b}\|} + \epsilon + \gamma \cdot \xi
\end{equation}
\( \beta \) scales the normalized gradient (ensuring it remains in a valid range). \( \epsilon \) prevents numerical instability. \( \gamma \) introduces random noise \( \xi \sim \mathcal{N}(0, \sigma^2) \) to further disrupt attacker optimization.

Patch-based adversarial attacks require a consistent gradient flow to optimize the perturbation across multiple training steps.
By randomizing gradients, we destabilize adversarial optimization, making it harder for patches to remain effective. Without GIR, a model trained with QAT could learn bit-width-specific gradient patterns. This makes it vulnerable to adaptive patch attacks that exploit predictable quantized gradients. By injecting gradient inconsistencies, the model learns more robust representations that generalize across multiple bit-widths.

Patch-based attacks often maintain their effectiveness across different bit-widths because the gradients remain highly structured.
GIR forces the model to learn irregular, bit-width-agnostic features, reducing the attack success rate when patches are transferred to new bit-widths.

\section{Results and Analysis}
\label{results}

\subsection{Experimental Setup}
\noindent\textbf{Dataset:} We evaluate our methods on two benchmark datasets: CIFAR-10 \cite{cifar} and ImageNet \cite{krizhevsky2017imagenet}. CIFAR-10 is used for controlled experimentation on smaller models, while ImageNet provides a more challenging and diverse evaluation setting for large-scale architectures.

\noindent\textbf{Model Architectures:}
 Our experiments span a variety of architectures, covering both convolutional neural networks (CNNs) and vision transformers (ViTs). 
ResNet: ResNet-56, ResNet-34, ResNet-20, ResNet-18 \cite{he2016deep}.
VGG: VGG-19, VGG-16 \cite{simonyan2014very}.
Other CNNs: AlexNet \cite{krizhevsky2017imagenet}, Inception-v3 \cite{szegedy2016rethinking}, Densenet-121 \cite{huang2018dense}.
Vision Transformers (ViTs): Swin-S \cite{liu2021swin}, DeiT-B \cite{deit}.

\noindent\textbf{Patch-Based Attacks:}
 To assess model vulnerability to structured adversarial perturbations, we employ three state-of-the-art patch-based attack methods:
LAVAN \cite{lavan}, 
Adversarial Patch GoogleAP (GAP) \cite{googleap}, 
Deformable Patch Representation (DPR) \cite{chen2022shape}. 

\noindent\textbf{Pixel-Level Attacks:}
 For evaluating traditional adversarial robustness, we consider four widely used pixel-level attack methods:
Fast Gradient Sign Method (FGSM) \cite{goodfellow2014explaining}, 
Projected Gradient Descent (PGD) \cite{madry2017towards}, 
Momentum Iterative Fast Gradient Sign Method (MIFGSM) \cite{dong2018boosting}, 
AutoAttack (AA) \cite{croce2020reliable}. 

Further experimental details including hyperparameter settings and implementation specifics are provided in the supplementary material.
\subsection{QADT-R: Experimental Results}

\subsubsection{Effect of QADT-R on Clean Accuracy Across Bit-Widths}

While improving adversarial robustness, a key objective of QADT-R is to maintain high clean accuracy. We measure the clean accuracy of models trained with QADT-R and compare them to baseline QAT and PBAT models across different quantization levels. 

\begin{table}[ht]
    \centering
    \footnotesize  
    \renewcommand{\arraystretch}{0.9}  
    \setlength{\tabcolsep}{4pt}  
    \begin{tabular}{|c|c|c|c|c|c|}
        \hline
        \textbf{Defense} & \textbf{32-bit} & \textbf{8-bit} & \textbf{5-bit} & \textbf{4-bit} & \textbf{2-bit} \\ \hline
        Standard QAT  & 89.4  & 87.8  & 85.1  & 80.5  & 78.2 \\ \hline
        PBAT~\cite{rao2020adversarial}  & 88.2  & 84.6  & 81.5  & 77.8  & 75.5 \\ \hline
        \textbf{QADT-R (Ours)}  & \textbf{87.9}  & \textbf{84.3}  & \textbf{81.2}  & \textbf{77.2}  & \textbf{75.1} \\ \hline
    \end{tabular}
    \caption{Clean accuracy (\%) of ResNet-54 (CIFAR-10) trained with Standard QAT, PBAT, and QADT-R across different bit-widths.}
    \label{tab:qadt_r_clean_acc_cifar}
\end{table}

As illustrated in Tables \ref{tab:qadt_r_clean_acc_cifar} and \ref{tab:qadt_r_clean_acc_imagenet}, QADT-R maintains competitive clean accuracy, with only a slight reduction compared to standard QAT, primarily due to the incorporation of adversarial training. However, this reduction is minimal, demonstrating that QADT-R effectively enhances robustness without significantly degrading standard model performance. In particular, compared to PBAT, QADT-R achieves slightly better clean accuracy, indicating that the dynamic bit-width training strategy does not overly constrain the model’s expressiveness. While lower-bit models naturally experience some accuracy degradation, as expected in QNNs, QADT-R remains closely aligned with baseline QAT models, ensuring a balanced trade-off between robustness and performance.

\begin{table}[ht] 
\centering 
\footnotesize
\renewcommand{\arraystretch}{0.9}
\setlength{\tabcolsep}{4pt}
\begin{tabular}{|c|c|c|c|c|c|} 
\hline 
\textbf{Defense} & \textbf{32-bit} & \textbf{8-bit} & \textbf{5-bit} & \textbf{4-bit} & \textbf{2-bit} \\ \hline 
Standard QAT & 85.2 & 82.8 & 80.1 & 77.5 & 73.9 \\ \hline PBAT~\cite{rao2020adversarial} & 84.1 & 80.9 & 78.2 & 75.4 & 71.8 \\ \hline 
\textbf{QADT-R (Ours)} & \textbf{84.5} & \textbf{81.3} & \textbf{78.7} & \textbf{75.9} & \textbf{72.3} \\ \hline 
\end{tabular} 
\caption{Clean accuracy (\%) of ResNet-34 (ImageNet) trained with Standard QAT, PBAT, and QADT-R across different bit-widths.} \label{tab:qadt_r_clean_acc_imagenet} 
\end{table}

\subsubsection{Effect of Bit-Width on Adversarial Robustness}

To systematically evaluate the impact of QADT-R across different quantization levels, we conduct experiments at bit-widths 32-bit, 8-bit, 5-bit, 4-bit, and 2-bit. We measure ASR reduction for each setting, comparing QADT-R against PBAT and DWQ. To further assess generalization, we introduce unseen adversarial patches, which are generated with different patch structures, scales, and quantization parameters not observed during training.

\begin{table}[ht]
    \centering
    \footnotesize  
    \renewcommand{\arraystretch}{0.9}  
    \setlength{\tabcolsep}{4pt}  
    \begin{tabular}{|c|c|c|c|c|c|}
        \hline
        \textbf{Defense} & \textbf{32-bit} & \textbf{8-bit} & \textbf{5-bit} & \textbf{4-bit} & \textbf{2-bit} \\ \hline
        PBAT~\cite{rao2020adversarial}  & 51.4  & 46.7  & 45.8  & 43.2  & 39.7 \\ \hline
        DWQ~\cite{fu2021double}         & 87.9  & 82.4  & 79.7  & 77.2  & 76.4 \\ \hline
        \textbf{QADT-R (Ours)}         & \textbf{28.5}  & \textbf{24.2}  & \textbf{22.8}  & \textbf{20.5}  & \textbf{19.3} \\ \hline
        PBAT (Unseen patch)  & 77.2  & 73.7  & 71.4  & 67.8  & 65.3 \\ \hline
        DWQ (Unseen patch)   & 87.8  & 82.6  & 79.2  & 77.9  & 76.3 \\ \hline
        \textbf{QADT-R (Unseen patch)}  & \textbf{30.1}  & \textbf{26.7}  & \textbf{24.9}  & \textbf{22.3}  & \textbf{21.0} \\ \hline
    \end{tabular}
    \caption{ASR (\%) of LAVAN attack against ResNet-54 (CIFAR-10) for different bit-width settings using QADT-R, PBAT, and DWQ, including results for unseen patches.}
    \label{tab:qadt_r_bitwidth_cifar_unseen}
\end{table}

QADT-R consistently demonstrates high effectiveness across all quantization levels, achieving significant reductions in ASR, even at extreme quantization settings like 2-bit precision, where it outperforms PBAT and DWQ by over 20\% and 50\%, respectively. When evaluating unseen patches, PBAT experience substantial ASR increases (25\%), confirming their overfitting to specific patch configurations. In contrast, QADT-R maintains strong robustness, with only a minor ASR increase of 1.6\%, demonstrating its ability to generalize effectively against new, unseen adversarial patch structures. DWQ is ineffective against patch-based attacks. 

\begin{table}[ht]
    \centering
    \footnotesize  
    \renewcommand{\arraystretch}{0.9}  
    \setlength{\tabcolsep}{4pt}  
    \begin{tabular}{|c|c|c|c|c|c|}
        \hline
        \textbf{Defense} & \textbf{32-bit} & \textbf{8-bit} & \textbf{5-bit} & \textbf{4-bit} & \textbf{2-bit} \\ \hline
        PBAT~\cite{rao2020adversarial}  & 51.4  & 47.2  & 43.9  & 40.8  & 37.1 \\ \hline
        DWQ~\cite{fu2021double}  & 86.7  & 71.3  & 67.8  & 63.5  & 61.2 \\ \hline
        \textbf{QADT-R (Ours)}  & \textbf{33.1}  & \textbf{28.4}  & \textbf{25.7}  & \textbf{23.5}  & \textbf{21.8} \\ \hline
        PBAT (Unseen)  & 78.5  & 73.7  & 70.1  & 67.2  & 64.6 \\ \hline
        DWQ (Unseen)  & 86.9  & 72.3  & 66.8  & 62.4  & 60.2 \\ \hline
        \textbf{QADT-R (Unseen)}  & \textbf{35.4}  & \textbf{31.6}  & \textbf{27.9}  & \textbf{25.6}  & \textbf{23.7} \\ \hline
    \end{tabular}
    \caption{ASR (\%) of LAVAN attack against ResNet-34 (ImageNet) for different bit-width settings using QADT-R, PBAT, and DWQ, including results for unseen patches.}
    \label{tab:qadt_r_bitwidth_imagenet_unseen}
\end{table}

On ImageNet, QADT-R consistently maintains low ASR values across all bit-width settings, significantly outperforming PBAT and DWQ in mitigating patch-based adversarial threats. PBAT exhibit ASR increases of up to 27\% when tested on unseen patches, indicating their failure to adapt to new attack configurations. In contrast, QADT-R experiences only a minor ASR increase of 2.3\%, demonstrating its strong generalization ability to novel adversarial patterns.

At 2-bit precision, QADT-R achieves the lowest ASR (23.7\%), confirming that its defenses remain highly effective even under extreme quantization conditions. This highlights QADT-R’s ability to counter adversarial patch transferability, ensuring robust protection across diverse quantization settings and real-world deployment scenarios.
\begin{table}[h]
    \centering
    \footnotesize  
    \renewcommand{\arraystretch}{0.85}  
    \setlength{\tabcolsep}{4pt}  
    \begin{tabular}{|c|c|c|c|c|c|}
    \hline
    \textbf{Defense Configuration} & \textbf{32-bit} & \textbf{8-bit} & \textbf{5-bit} & \textbf{4-bit} & \textbf{2-bit} \\ \hline
    QADT-R without A-QAPA & 34.5 & 36.8 & 37.4 & 38.2 & 39.5 \\ \hline
    QADT-R without DBWT & 30.2 & 31.9 & 33.5 & 34.8 & 36.1 \\ \hline
    QADT-R without GIR & 27.6 & 29.4 & 30.1 & 31.7 & 33.2 \\ \hline
    QADT-R Full & \textbf{24.8} & \textbf{26.5} & \textbf{28.2} & \textbf{29.5} & \textbf{31.3} \\ \hline
    \end{tabular}
    \caption{ASR (\%) across different bit-widths for CIFAR-10.}
    \label{tab:ablation_qadt_r_bitwidth_cifar}
\end{table}
\subsubsection{Ablation Study: Impact of QADT-R Components}
The ablation study highlights the importance of each component in QADT-R by analyzing its impact on adversarial robustness. A-QAPA plays a crucial role in mitigating patch transferability across bit-widths. When A-QAPA is removed, the ASR increases significantly, indicating that adversarial patches generated without simulated quantization remain highly effective across different bit-widths. DBWT also contributes substantially to the model’s robustness. Without DBWT, the model overfits to a specific bit-width, failing to generalize defenses across varying quantization settings, resulting in a higher ASR. The most pronounced drop in robustness occurs when GIR is disabled. This confirms that GIR effectively disrupts adversarial optimization by introducing gradient perturbations, making it harder for the attack to align across bit-widths. Finally, the complete QADT-R framework achieves the lowest ASR, demonstrating that all three components work synergistically to enhance adversarial robustness.
\begin{table}[h]
    \centering
    \footnotesize  
    \renewcommand{\arraystretch}{0.85}  
    \setlength{\tabcolsep}{4pt}  
    \begin{tabular}{|c|c|c|c|c|c|}
    \hline
    \textbf{Defense Configuration} & \textbf{32-bit} & \textbf{8-bit} & \textbf{5-bit} & \textbf{4-bit} & \textbf{2-bit} \\ \hline
    QADT-R without A-QAPA & 38.7 & 40.2 & 41.1 & 42.3 & 44.1 \\ \hline
    QADT-R without DBWT & 35.1 & 37.0 & 38.3 & 39.7 & 41.4 \\ \hline
    QADT-R without GIR & 32.5 & 34.2 & 35.5 & 36.8 & 38.6 \\ \hline
    QADT-R Full & \textbf{28.5} & \textbf{30.1} & \textbf{31.7} & \textbf{33.1} & \textbf{35.0} \\ \hline
    \end{tabular}
    \caption{ASR (\%) across different bit-widths for ImageNet.}
    \label{tab:ablation_qadt_r_bitwidth_imagenet}
\end{table}

The results of ablation study in Table \ref{tab:ablation_qadt_r_unseen_cifar} and Table \ref{tab:ablation_qadt_r_unseen_imagenet} demonstrate the contribution of each component in QADT-R when tested against unseen patch configurations. The removal of A-QAPA results in the largest increase in ASR, with ASR rising by 11.8\% at 2-bit on CIFAR-10 and 11.2\% on ImageNet. This highlights the critical role of A-QAPA in mitigating transferable adversarial patches by ensuring patches are optimized within the quantized domain during attack generation. Without this step, patches remain highly effective across bit-widths.
\begin{table}[h]
    \centering
    \footnotesize  
    \renewcommand{\arraystretch}{0.85}  
    \setlength{\tabcolsep}{4pt}  
    \begin{tabular}{|c|c|c|c|c|c|}
    \hline
    \textbf{Defense Configuration} & \textbf{32-bit} & \textbf{8-bit} & \textbf{5-bit} & \textbf{4-bit} & \textbf{2-bit} \\ \hline
    QADT-R without A-QAPA & 38.2 & 41.1 & 42.3 & 43.7 & 45.9 \\ \hline
    QADT-R without DBWT & 34.5 & 36.9 & 38.6 & 40.1 & 42.3 \\ \hline
    QADT-R without GIR & 31.8 & 33.7 & 35.2 & 36.9 & 38.8 \\ \hline
    QADT-R Full & \textbf{26.4} & \textbf{28.5} & \textbf{30.2} & \textbf{32.1} & \textbf{33.9} \\ \hline
    \end{tabular}
    \caption{ASR (\%) across different bit-widths for CIFAR-10 with unseen patches.}
    \label{tab:ablation_qadt_r_unseen_cifar}
\end{table}
Similarly, removing DBWT significantly weakens robustness, leading to an ASR increase of 8.4\% at 4-bit on CIFAR-10 and 7.9\% at 5-bit on ImageNet. This confirms that fixed-bit adversarial training results in overfitting, making models highly vulnerable when evaluated at unseen bit-widths. The flexibility of DBWT ensures that models trained under varying precision settings generalize well to real-world mixed-precision inference scenarios.

Furthermore, GIR remains effective across both datasets. Even for previously unseen patches, removing GIR increases ASR by 6.7\% at 8-bit on CIFAR-10 and 5.9\% on ImageNet, demonstrating its role in disrupting adversarial optimization. GIR prevents gradient-aligned learning, making it harder for adversaries to find stable, transferable perturbations.
\begin{table}[h]
    \centering
    \footnotesize  
    \renewcommand{\arraystretch}{0.85}  
    \setlength{\tabcolsep}{4pt}  
    \begin{tabular}{|c|c|c|c|c|c|}
    \hline
    \textbf{Defense Configuration} & \textbf{32-bit} & \textbf{8-bit} & \textbf{5-bit} & \textbf{4-bit} & \textbf{2-bit} \\ \hline
    QADT-R without A-QAPA & 42.1 & 44.3 & 45.7 & 47.0 & 49.1 \\ \hline
    QADT-R without DBWT & 38.5 & 40.9 & 42.4 & 44.1 & 46.2 \\ \hline
    QADT-R without GIR & 35.6 & 37.5 & 39.2 & 40.8 & 42.6 \\ \hline
    QADT-R Full & \textbf{30.1} & \textbf{32.7} & \textbf{34.5} & \textbf{36.0} & \textbf{37.9} \\ \hline
    \end{tabular}
    \caption{ASR (\%) across different bit-widths for ImageNet with unseen patches.}
    \label{tab:ablation_qadt_r_unseen_imagenet}
\end{table}
Finally, the full QADT-R framework achieves the lowest ASR across all bit-widths, validating the synergistic effect of A-QAPA, DBWT, and GIR in reducing adversarial transferability. 

\section{Conclusion}
\label{conclusion}

This work demonstrates that patch-based adversarial attacks remain highly transferable across QNNs, maintaining high attack success rates despite reduced precision. To address this vulnerability, we proposed Quantization-Aware Defense Training with Randomization (QADT-R), integrating Adaptive Quantization-Aware Patch Generation (A-QAPA), Dynamic Bit-Width Training (DBWT), and Gradient-Inconsistent Regularization (GIR). Our approach effectively reduces ASR by up to 25\%-40\% compared to prior defenses (PBAT, DWQ) while maintaining competitive clean accuracy. 
{
    \small
    \bibliographystyle{ieeenat_fullname}
    \bibliography{main}
}

\clearpage
\setcounter{page}{1}
\maketitlesupplementary

\subsection{Background}
\label{background}
This section provides a detailed overview of key concepts and techniques relevant to the study, including quantization in neural networks, pixel-level and patch-based adversarial attacks, and transferability challenges. It serves as a foundation for understanding the experiments and analyses presented in both the main paper and supplementary material.
\subsubsection{Quantization in Neural Networks}
Deep neural network (DNN) quantization compresses and accelerates models by representing weights, activations, and sometimes gradients with lower bit widths. Quantization is a cornerstone of efficient DNN inference frameworks in industry, enabling deployment on resource-constrained devices such as edge hardware and mobile platforms \cite{jacob2018quantization}. There are two main categories of quantization techniques: Quantization-Aware Training (QAT) \cite{hubara2016binarized, rastegari2016xnor, zhou2016dorefa, li2019additive} and Post-Training Quantization (PTQ) \cite{nagel2020up, li2021brecq, hubara2021accurate, wei2022qdrop}.

\noindent\textbf{Quantization-Aware Training (QAT):} QAT integrates quantization effects into the training process, either by training DNNs from scratch or by fine-tuning pre-trained full-precision models. Given a training dataset $D_{train} = \{x_i,y_i\}^n_{i=1}$, QAT aims to optimize the weights $\omega$ while adapting the model to quantization-induced noise. This approach typically achieves lower quantization loss compared to PTQ but requires access to the full training dataset and additional computational resources for end-to-end training.

The QAT process is often implemented using fake quantization, where weights remain stored in full precision but are rounded to lower bit-width values during inference. The training objective for QAT with fake quantization can be formalized as:

\begin{equation}
    \min_{\omega,\beta} \sum^n_{i=1} L(f(x_i,\omega,\beta), y_i),
\end{equation}

where $L(\cdot)$ is the loss function (e.g., cross-entropy loss), $\omega$ represents the model weights, and $\beta$ denotes the quantization hyperparameter used to map full-precision weights to lower bit-width representations.

\noindent\textbf{Post-Training Quantization (PTQ):} PTQ offers a computationally efficient alternative to QAT, requiring no end-to-end retraining. Instead, PTQ relies on a smaller calibration dataset $D_{cali} = \{x_i,y_i\}^m_{i=1}$, where $m << n$,  to optimize quantization parameters $\beta$. Model weights $\omega$ remain unchanged during PTQ, focusing exclusively on minimizing quantization-induced errors. This process can also be implemented with fake quantization and is formalized as:
\begin{equation}
    \min_{\beta} \sum^n_{j=1} L(f(x_j,\omega,\beta), y_j),
\end{equation}
While PTQ is computationally lightweight, it often incurs higher quantization loss compared to QAT, particularly at lower bit widths.

Recent advancements in PTQ \cite{nagel2020up, li2021brecq, hubara2021accurate, wei2022qdrop} have modeled weight and activation quantization as perturbation problems, using Taylor expansion to analyze loss value changes and reconstruct outputs for individual layers. These methods have significantly improved the accuracy of PTQ models, even under aggressive quantization constraints.

\noindent\textbf{Training Details:}
For CIFAR-10 dataset, all models were trained with a batch size of 128 using Stochastic Gradient Descent (SGD) with a momentum of 0.9 and weight decay set to 
$1\times10^{-4}$. Each model was trained for 50 epochs to ensure adequate convergence across quantization levels and architectures.

For the ImageNet dataset, we follow the approach of previous works \cite{wang2021feature, yang2024quantization}, conducting experiments on a curated subset of the ImageNet 2012 validation set, which consists of 1,000 high-resolution images (299×299 pixels). 
\subsubsection{Pixel-Level Adversarial Attacks }
Adversarial attacks involve adding small, carefully crafted perturbations to input data to manipulate a model’s predictions. These perturbations are often imperceptible to humans but are sufficient to mislead the model into making incorrect classifications. Common types of adversarial attacks include:


\noindent \textbf{Fast gradient sign method (FGSM)} \cite{goodfellow2014explaining} is a single-step, gradient-based, attack. An adversarial example is generated by performing a one step gradient update along the direction of the sign of gradient at each pixel as follows:

 \begin{equation}
     x^{adv} = x - \epsilon \cdot sign (\nabla_{x}J(x,y))
 \end{equation}
Where $\nabla J()$ computes the gradient of the loss function $J$ and $\theta$ is the set of model parameters. The $sign()$ denotes the sign function and $\epsilon$ is the perturbation magnitude. 

\noindent \textbf{Projected Gradient Descent (PGD)} \cite{madry2017towards} is an iterative variant of the FGSM where the adversarial example is generated as follows:
 \begin{equation}
x^{t+1} = \mathcal{P}_{\mathcal{S}_x}(x^t + \alpha \cdot sign (\nabla_{x}\mathcal{L}_{\theta}(x^t,y)) )
 \end{equation}
Where $\mathcal{P}_{\mathcal{S}_x}()$ is a projection operator projecting the input into the feasible region $\mathcal{S}_x$ and $\alpha$ is the added noise at each iteration. The PGD attack tries to find the perturbation that maximizes the loss on a particular input while keeping the size of the perturbation smaller than a specified amount. 

\noindent \textbf{Momentum Iterative Fast Gradient Sign Method (MIFGSM)} \cite{dong2018boosting} introduced a momentum term to stabilize the update direction during the iteration.
 \begin{equation}
     g_{t+1} = \mu \cdot g_{t} + \frac{\nabla_{x}J(x_{t+1}^{adv},y)}{\parallel \nabla_{x}J(x_{t+1}^{adv},y) \parallel_1}
 \end{equation}
 \begin{equation}
     x_{t+1}^{adv} = x_{t}^{adv} - \alpha \cdot sign (g_{t+1})
 \end{equation}

\noindent \textbf{Autoattack (AA)} \cite{croce2020reliable} is a strong, automated adversarial attack framework designed to provide a reliable and robust evaluation of adversarial defenses. It is widely used in adversarial robustness benchmarks due to its effectiveness and parameter-free nature. AutoAttack is not a single attack, but rather a combination of four diverse attacks that complement each other: APGD-CE (Auto-PGD with Cross-Entropy Loss): A strong white-box attack optimized using Projected Gradient Descent (PGD).
APGD-DLR (Auto-PGD with Difference of Logits Ratio Loss): A variant of APGD that improves success on robust models.
FAB (Fast Adaptive Boundary Attack): A decision-based attack that efficiently finds minimal adversarial perturbations.
Square Attack: A query-efficient black-box attack based on random search.



\subsubsection{Patch-based Adversarial Attacks} 
Patch-based adversarial attacks introduce highly visible, localized perturbations that redirect the model's focus, often achieving high success rates with minimal modifications \cite{guesmi2023physical}. Unlike pixel-level attacks, which subtly perturb individual pixel values across the entire input, patch-based attacks concentrate their impact within a small, localized region. This focused approach not only makes them more resilient to noise and transformations but also enhances their applicability in real-world, physically realizable scenarios. Patch attacks have been successfully demonstrated across various domains, including image classification \cite{lavan, googleap, li2021generative, chen2022shape} and object detection \cite{guesmi2024dap}, showcasing their high transferability across diverse model architectures.

Despite their demonstrated effectiveness and unique threat potential, the interaction between patch-based attacks and QNNs remains underexplored. Specifically, there is limited research on how quantization impacts the robustness and transferability of patch-based attacks, leaving a significant gap in understanding their efficacy against QNNs deployed in resource-constrained environments.

\textbf{Techniques for Patch-Based Adversarial Attacks:}

\begin{itemize}
    \item \textbf{Localized and Visible Adversarial Noise (LAVAN):} The LAVAN technique \cite{lavan} generates adversarial patches that are both localized and highly visible. These patches are trained iteratively by randomly selecting images and placing the patch at varying locations. This randomization ensures that the patch generalizes well across different contexts and positions, capturing distinctive adversarial features that make it both highly transferable and effective. By applying the patch to diverse scenarios, LAVAN achieves robustness to environmental variations, making it a strong candidate for both digital and physical adversarial settings.
    \item \textbf{Google Adversarial Patch (GAP):} GAP attack \cite{googleap} emphasizes practicality in real-world attacks, addressing challenges associated with traditional \( L_p \)-norm-based adversarial perturbations, such as requiring precise object capture through a camera. GAP creates universal adversarial patches that can be applied to any part of an input, making it more versatile. Furthermore, it incorporates the Expectation over Transformation (EOT) \cite{eot} framework to improve the patch’s robustness against common variations like rotation, scaling, and translation, ensuring its effectiveness under real-world conditions.
    \item \textbf{Deformable Patch Representation (DPR):} DRP attack \cite{chen2022shape} is a type of adversarial attack that focuses on altering the shape and appearance of patches in an image rather than just modifying pixel values. The main idea is to create deformable and dynamic patches that can adapt their form to exploit the vulnerabilities of neural networks more effectively. DPR essentially aims to increase the robustness and transferability of adversarial patches by introducing shape-based deformations, which makes the attack less dependent on specific pixel perturbations and more adaptable across different models and input variations.
\end{itemize}

\subsection{Experimental Setup}

Table \ref{tab:hyperparameter} presents the set of hyperparameter configurations used in our experiments for adversarial patch generation, dynamic bit-width training, and gradient-inconsistent regularization, ensuring robustness against patch-based attacks in quantized neural networks.

\begin{table*}[h]
    \centering
    \begin{tabular}{|c|c|c|c|}
        \hline
        \textbf{Hyperparameter} & \textbf{Symbol} & \textbf{Description} & \textbf{Values} \\
        \hline
        Gradient Perturbation Scale   & $\alpha$   & Strength of perturbation added to gradients   & $0.1$   \\
        Randomization Factor   & $\beta$  & Scaling of normalized gradients   & $1.0$  \\
        Noise Standard Deviation  & $\sigma$  & Standard deviation of random noise applied to gradients & $0.05$  \\
        Numerical Stability Term & $\epsilon$  & Small constant to prevent instability  & $10^{-5}$  \\
        Gradient Disruption Frequency & $f_g$  & Fraction of training steps where perturbations are applied & $50\%$  \\
        Weight Decay & $\lambda$ & Regularization factor for preventing overfitting to bit-widths& $10^{-4}$ \\
        Batch Size & $N$ & Number of samples per batch & $128$ \\
        Quantization Noise & $\sigma_b$ & Variance of bit-width switching noise & 0.01 \\
        Epoch Range per Bit-Width & $E_b$ & Number of epochs per bit-width level & 20 \\
        \hline
    \end{tabular}
    \caption{Hyperparameter configurations used in our experiments for adversarial patch generation, dynamic bit-width training, and gradient-inconsistent regularization, ensuring robustness against patch-based attacks in quantized neural networks.}
    \label{tab:hyperparameter}
\end{table*}
\subsection{Why Patch-Based Attacks Transfer Well?}
\label{sec:why}


\subsubsection{Additional Results: Effect of Quantization Levels on Patch Transferability }
We evaluate the robustness of Inception v3 and DenseNet-121 under patch-based adversarial attacks at different quantization bit-widths. As shown in Table~\ref{tab:experiment2_inception}, the LAVAN attack remains highly effective even at lower bit-widths, demonstrating that quantization alone does not significantly reduce patch-based adversarial effectiveness. Notably:

At 2-bit precision, Inception v3 and DenseNet-121 retain over 50\% attack success rates (ASR), highlighting the persistent vulnerability of quantized models to structured adversarial patches.
Lowering the bit-width reduces ASR slightly, but the overall effectiveness of LAVAN remains high, particularly at 4-bit and 8-bit precision.
This confirms that spatially constrained perturbations (patches) survive quantization effects, unlike pixel-wise perturbations, which suffer from gradient masking.

\begin{table}[ht]
    \centering
    \footnotesize  
    \renewcommand{\arraystretch}{0.9} 
    \setlength{\tabcolsep}{4pt} 
    \begin{tabular}{|c|c|c|c|c|}
        \hline
        \textbf{Model} & \textbf{32-bit} & \textbf{8-bit} & \textbf{4-bit} & \textbf{2-bit} \\ \hline
        Inception v3  & 89.10 & 60.12  & 55.32 & 50.66 \\ \hline
        DenseNet-121  & 88.35 & 59.43  & 54.97 & 49.25 \\ \hline        
    \end{tabular}
    \caption{ASR (\%) of LAVAN attack with a noise percentage 0.08 on ImageNet. }
    \label{tab:experiment2_inception}
\end{table}

In addition to LAVAN, we evaluate another state-of-the-art patch-based attack, Deformable Patch Representation (DPR) \cite{chen2022shape}, against ResNet34 on ImageNet. As shown in Table~\ref{tab:experiment2_DPR_resnet34}, DPR remains highly effective across all bit-widths, achieving a 49.87\% ASR even at 2-bit precision.

DPR maintains high attack success rates, similar to LAVAN, reinforcing the transferability of patch-based attacks in QNNs.
Even at 4-bit and 2-bit, the model remains highly vulnerable, confirming that quantization does not provide significant protection against structured attacks.
Unlike pixel-level adversarial attacks, DPR does not rely on fine-grained gradient signals, making it more resistant to quantization-induced robustness artifacts.

\begin{table}[ht]
    \centering
    \footnotesize  
    \renewcommand{\arraystretch}{0.9}  
    \setlength{\tabcolsep}{4pt}  
    \begin{tabular}{|c|c|c|c|c|}
        \hline
        \textbf{Attack}  & \textbf{32-bit} & \textbf{8-bit} & \textbf{4-bit} & \textbf{2-bit} \\ \hline
        DPR    & 83.46 & 59.32 & 55.82 & 49.87 \\ \hline
    \end{tabular}
    \caption{ASR (\%) of DPR attack against ResNet34 on ImageNet. }
    \label{tab:experiment2_DPR_resnet34}
\end{table}

\subsubsection{Why Patches Transfer Well? Feature Map Analysis Across Bit Widths}
\label{sec:why1}
\begin{figure*}
    \centering
    \includegraphics[width=0.8\linewidth]{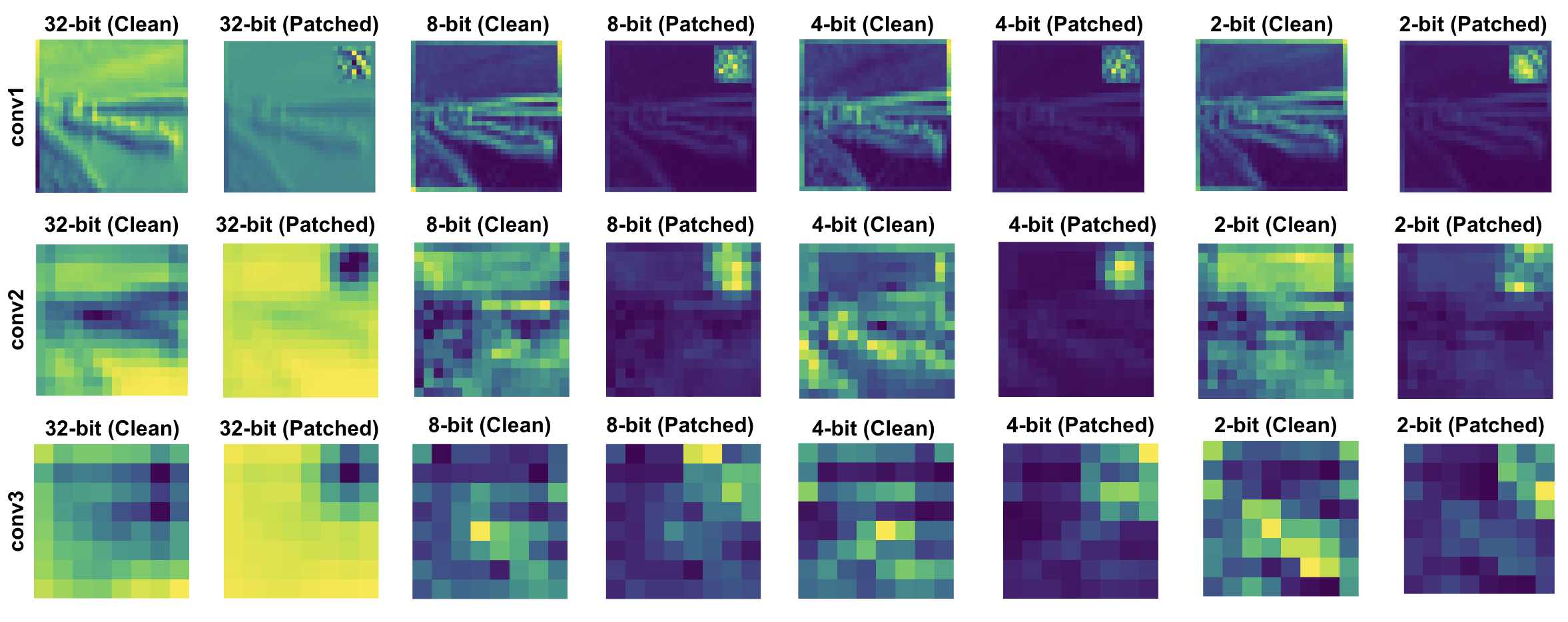}
    \caption{Feature maps of the 32-bit, 8-bit, 4-bit, and 2-bit models comparing the clean and patched feature maps for the three first convolutional layers.}
    \label{fig:experiment1}
\end{figure*}
To understand why patch-based adversarial attacks transfer effectively across quantized models, we analyze feature maps extracted from models quantized at different bit widths (32-bit, 8-bit, 4-bit, 2-bit) using QAT. We pass clean and patched inputs through these models and extract intermediate feature representations from key convolutional layers. These feature maps provide insights into how adversarial patches interact with network activations across quantization levels.

As shown in Figure \ref{fig:experiment1}, the patched feature maps exhibit strong localized activations at the patch location, regardless of the quantization level.
These high-intensity activations persist even at 2-bit quantization, suggesting that the patch effectively disrupts local feature representations in a way that survives precision reduction.

In each examined convolutional layer, the patch creates a structured, high-contrast response, unlike subtle pixel-wise adversarial perturbations that are often suppressed by quantization effects.
The visual distinction of the patch remains across all quantization settings, reinforcing its resilience to bit-width reduction.

While quantization slightly reduces feature intensity, it does not eliminate the patch’s impact.
Even in low-bit (4-bit, 2-bit) models, patch-based disruptions remain distinguishable, indicating that adversarial patches exploit robust, high-level features that quantization does not easily neutralize.

Unlike pixel-based adversarial attacks that rely on subtle, dispersed perturbations, patches leverage strong, localized patterns that remain effective even when model precision is reduced.
This explains why patch-based attacks exhibit high transferability across different quantization levels, as previously observed in our attack success rate experiments.
\subsubsection{Why Patches Transfer Well? Gradient Alignment Across Bit Widths for Patch vs. Pixel-Level Attacks} 
\label{sec:why2}
To analyze the transferability of patch-based adversarial attacks compared to pixel-level perturbations, we evaluate gradient alignment across different quantization levels. Specifically, we:

Generate adversarial gradients for patch-based (LAVAN) and pixel-level (PGD) attacks on a 32-bit full-precision model.
Compute Cosine Similarity between these gradients and those of quantized models (8-bit, 4-bit, 2-bit) trained with QAT to measure alignment.
Measure the Mean Squared Error (MSE) between gradients to quantify the magnitude of gradient deviations caused by quantization.

\begin{figure*}
    \centering
    \includegraphics[width=0.9\linewidth]{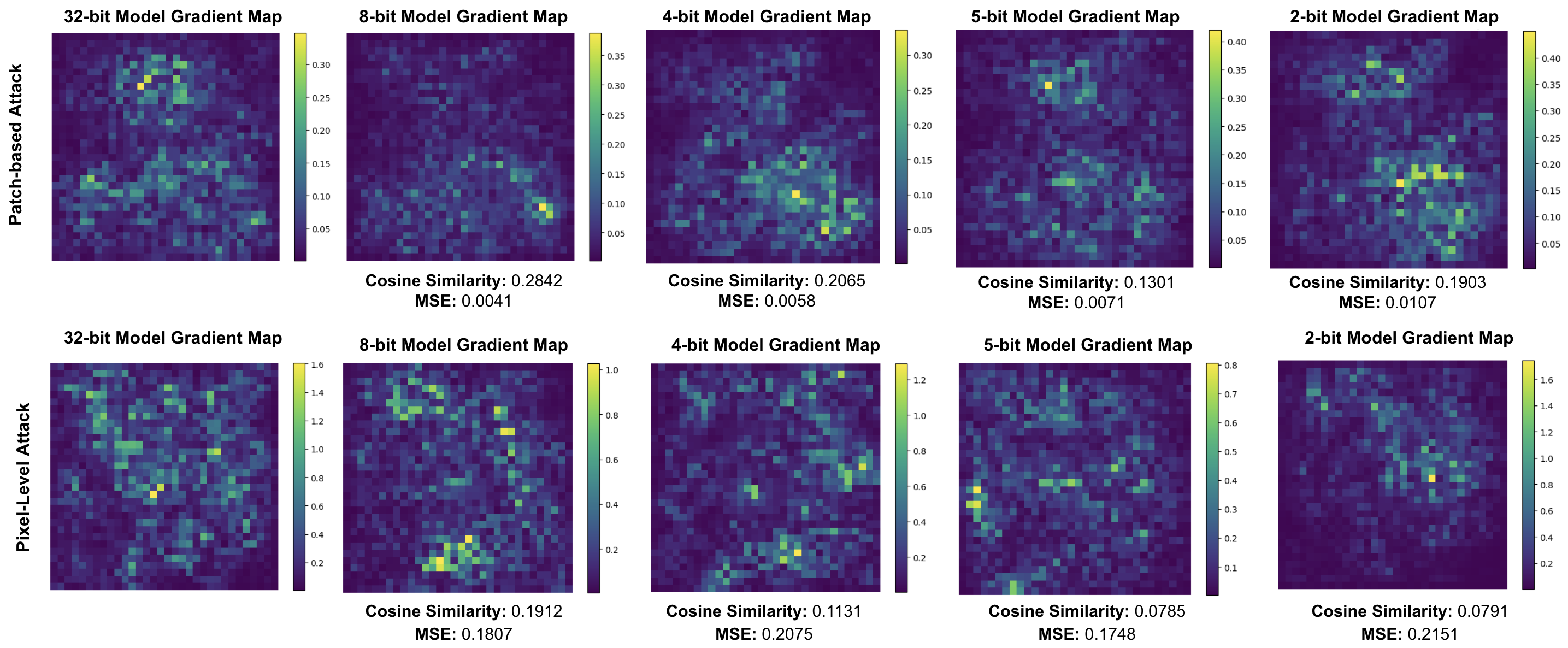}
    \caption{Gradient maps for 32-bit, 8-bit, 4-bit, and 2-bit models under patch-based and pixel-level attacks, along with Cosine Similarity and MSE measurements comparing gradients between the full-precision and quantized models.}
    \label{fig:experiment4}
\end{figure*}

As shown in Figure \ref{fig:experiment4}, the cosine similarity between patch attack gradients in 32-bit and quantized models remains relatively high across bit-widths (8-bit: 0.2842, 5-bit: 0.1301, 2-bit: 0.1026). In contrast, pixel-level attack gradients suffer from more severe misalignment, with cosine similarity dropping as low as 0.0791 at 2-bit, indicating a significant loss of structure due to quantization.

The MSE for patch-based attack gradients is consistently lower than that of pixel-based attacks, meaning that patch attack gradients preserve their magnitude more effectively across quantized models. For Patch-based attack, MSE is equal to (8-bit: 0.0041
2-bit: 0.0107) as for the pixel-level MSE is equal to (8-bit: 0.1807, 2-bit: 0.2151). These results confirm that quantization disrupts pixel-level attack gradients more than patch-based attacks. 

The gradient maps in Figure \ref{fig:experiment4} reveal that patch-based attacks preserve visually distinct activation regions across all quantization settings. However, pixel-level attack gradients become increasingly unstructured and diffused at lower bit-widths, reflecting the loss of fine-grained details due to quantization.

\subsubsection{Why Patches Transfer Well? Spatial Sensitivity of Patches} 
\label{sec:why3}
To further analyze the robustness of patch-based adversarial attacks in quantized models, we evaluate how spatial modifications (shifts and rotations) affect attack effectiveness across different bit-widths. 

Patches are generated on a 32-bit model and transferred to QAT-trained models quantized to 8-bit, 5-bit, 4-bit, and 2-bit. The attack success rate (ASR) is recorded for:
Shifting the patch by (2,2), (8,8), and (18,18) pixels.
Rotating the patch by 5°, 10°, 20°, and 30°

\begin{table}[ht]
    \centering
    \footnotesize  
    \renewcommand{\arraystretch}{0.9}  
    \setlength{\tabcolsep}{4pt}  
    \begin{tabular}{|c|c|c|c|c|c|}
    \hline
         &  \textbf{32-bit} & \textbf{8-bit} & \textbf{5-bit} & \textbf{4-bit} & \textbf{2-bit}\\
    \hline
      Original            & 86.43  & 83.24  & 79.94 & 76.22 & 73.08 \\ \hline
      Shifted (2,2)       & 87.73  & 82.35  & 79.55 & 77.39 & 72.93 \\ \hline
      Shifted (8,8)       & 85.13  & 82.56  & 80.24 & 78.11 & 72.88 \\ \hline
      Shifted (18,18)     & 82.47  & 80.77  & 79.04 & 74.78 & 70.10 \\ \hline
      Rotated $5^\circ$   & 86.39  & 82.98  &  80.81 & 76.16 & 72.84 \\ \hline
      Rotated $10^\circ$  & 86.24  & 82.68  &  80.45 & 76.79  & 72.20 \\ \hline
      Rotated $20^\circ$  & 84.38  & 81.50  &  79.85 & 75.17  & 71.73 \\ \hline
      Rotated $30^\circ$  & 82.98  & 80.71  &  78.19 & 74.82 & 70.39 \\
    \hline
    \end{tabular}
    \caption{ASR (\%) of LAVAN attack against ResNet-56 (full precision and quantized) on CIFAR-10 at various patch positions and rotation angle. }
    \label{tab:experiment3}
\end{table}

Minimal impact of small spatial shifts. As shown in Table \ref{tab:experiment3}, small displacements, such as (2,2) or (8,8), do not significantly degrade ASR, with success rates remaining above 70\% even at 2-bit quantization.
Larger shifts (18,18) result in a minor reduction in ASR, but patches remain highly effective across all bit-widths. 

Patch effectiveness is resilient to rotation. Even at 30° rotation, ASR remains above 80.71\% at 8-bit and 70.39\% at 2-bit, indicating strong persistence of adversarial features despite orientation changes.

Consistent transferability across quantized models. Despite increasing quantization constraints, patches retain strong adversarial impact, reinforcing their effectiveness as a highly transferable attack method.
Even at 2-bit precision, patches still achieve over 70\% ASR, demonstrating that quantization alone does not disrupt patch effectiveness.

\subsubsection{Why Patches Transfer Well? Impact of Patch Size and Visibility}  %
\label{sec:why4}
To analyze the effect of patch size on adversarial success, we generate patch-based attacks of different sizes on a 32-bit model and evaluate their attack success rates across quantization levels (8-bit, 5-bit, 4-bit, and 2-bit).

Larger patches increase attack success rates across all bit-widths, 12×12 patches achieve the highest ASR, exceeding 95.56\% on 32-bit and remaining effective even at 84.89\% on 2-bit models.
Smaller 6×6 patches experience a steeper ASR decline, suggesting that quantization affects smaller adversarial perturbations more than larger, high-visibility patches.

Consistent attack effectiveness despite quantization. While quantization reduces ASR overall, larger patches remain highly effective even at lower bit-widths (e.g., 12×12 patches still have an 84.89\% ASR at 2-bit).
This suggests that larger patches maintain stronger adversarial influence, counteracting quantization-induced distortions.

More visible patches exert stronger adversarial effects. Unlike pixel-level attacks, which degrade significantly under gradient masking, patch-based attacks rely on high-contrast, localized perturbations that remain effective even in low-bit QNNs.
The increased visibility of larger patches ensures that adversarial features persist across different quantization settings.
\begin{table}[ht]
    \centering
    \footnotesize  
    \renewcommand{\arraystretch}{0.9} 
    \setlength{\tabcolsep}{4pt} 
    \begin{tabular}{|c|c|c|c|c|c|}
    \hline
       \textbf{Size} &  \textbf{32-bit} & \textbf{8-bit} & \textbf{5-bit} & \textbf{4-bit} & \textbf{2-bit}\\
    \hline
      6x6   &  86.43  & 83.24  & 79.94 & 76.22 & 73.08\\ \hline
      8x8    & 89.92  & 86.94  & 81.62  & 79.76 & 75.18 \\ \hline
      10x10  & 91.92  & 88.94  & 83.50  & 81.16 & 78.62 \\ \hline
      12x12  & 95.56  & 90.10  & 88.52  & 86.19  & 84.89 \\ 
    \hline
    \end{tabular}
    \caption{ASR (\%) of LAVAN attack against ResNet-56 on CIFAR-10 for different patch sizes.}
    \label{tab:experiment6_resnet56}
\end{table}


\end{document}